\documentclass[conference]{IEEEtran}
\IEEEoverridecommandlockouts
\usepackage{cite}
\usepackage{amsmath,amssymb,amsfonts}
\usepackage{algorithmic}
\usepackage{graphicx}
\usepackage{textcomp}
\usepackage{url}
\usepackage{xcolor}
\usepackage{stfloats}
\usepackage{float}
\def\BibTeX{{\rm B\kern-.05em{\sc i\kern-.025em b}\kern-.08em
    T\kern-.1667em\lower.7ex\hbox{E}\kern-.125emX}}
\begin{document}

\title{Wavelet-Based GAN Fingerprint Detection using ResNet50}

\author{\IEEEauthorblockN{Sai Teja Erukude}
\IEEEauthorblockA{\textit{Department of Computer Science} \\
\textit{Kansas State University}\\
Manhattan, USA \\
erukude.saiteja@gmail.com}
\and
\IEEEauthorblockN{Suhasnadh Reddy Veluru}
\IEEEauthorblockA{\textit{College of Business Administration} \\
\textit{Kansas State University}\\
Manhattan, USA \\
suhasnadhreddyveluru@gmail.com}
\and
\IEEEauthorblockN{Viswa Chaitanya Marella}
\IEEEauthorblockA{\textit{College of Business Administration} \\
\textit{Kansas State University}\\
Manhattan, USA \\
viswachaitanyamarella@gmail.com}
}

\maketitle

\begin{abstract}
Identifying images generated by Generative Adversarial Networks (GANs) has become a significant challenge in digital image forensics. This research presents a wavelet-based detection method that uses discrete wavelet transform (DWT) preprocessing and a ResNet50 classification layer to differentiate the StyleGAN-generated images from real ones. Haar and Daubechies wavelet filters are applied to convert the input images into multi-resolution representations, which will then be fed to a ResNet50 network for classification, capitalizing on subtle artifacts left by the generative process. Moreover, the wavelet-based models are compared to an identical ResNet50 model trained on spatial data. The Haar and Daubechies preprocessed models achieved a greater accuracy of 93.8 percent and 95.1 percent, much higher than the model developed in the spatial domain (accuracy rate of 81.5 percent). The Daubechies-based model outperforms Haar, showing that adding layers of descriptive frequency patterns can lead to even greater distinguishing power. These results indicate that the GAN-generated images have unique wavelet-domain artifacts or ``fingerprints." The method proposed illustrates the effectiveness of wavelet-domain analysis to detect GAN images and emphasizes the potential of further developing the capabilities of future deepfake detection systems.
\end{abstract}


\section{Introduction and Related Work}\label{introduction}

Generative Adversarial Networks (GANs) can now produce extremely convincing visuals that are increasingly challenging for us to differentiate from actual photographs \cite{pativada2025gansvaes}. The wide dissemination of ``fake" images produced by GANs poses serious security and ethical risks, as people can be deceived by synthetic media \cite{mahara2025methodstrendsdetectinggenerated, tang2021detection}. In response, the field of digital image forensics has pivoted towards developing tools to detect AI-generated images with fidelity to promote trust in visual content. Despite significant progress, existing detection techniques struggle with generalization and robustness as generative models rapidly advance. Detecting GAN images remains an ongoing challenge, especially as new GAN variants minimize obvious artifacts that earlier detectors relied upon.

A variety of methods have been explored for GAN image detection. Early methods treated the problem as a binary classification of images, adopting deep convolutional networks on raw pixels. For instance, studies showed that CNN-synthesized visuals are relatively easy to identify when a ResNet classifier is trained to distinguish between GAN images and real images. Such pixel-domain CNN detectors successfully discriminated first-generation GANs with high accuracy. There are other examples of focusing on GAN anomalies in facial images, such as inconsistent eye reflections or misaligned features. For instance, researchers have applied facial landmarks to detect unnatural geometry or variations in head pose to identify deep fakes. However, as GAN architectures improve, these hand-crafted cues often lose reliability. GANs like StyleGAN2 can produce faces with correct geometry with fewer visible flaws, and reduce effectiveness for detectors that only consider spatial inconsistencies \cite{karras2020analyzingimprovingimagequality}.

One promising direction is to find the intrinsic fingerprints of the GAN generation process. Just as real cameras imprint sensor noise patterns (PRNU) on photographs, GANs also leave model-specific patterns as fingerprints of their generative processes. This insight paved the way for a new method of detection and attribution focused on fingerprints, and subsequent work has attempted to apply spectral or frequency analysis to recover fingerprints. For example, some methods transform images to the frequency space, where differences between real and generated images become more pronounced, and then apply a classifier. Studies suggest that frequency analysis can amplify subtle artifacts that are less visible in the pixel domain, thereby aiding detection \cite{computers13120341, erukude2024identifying, mahara2025methodstrendsdetectinggenerated, zhang2019detectingsimulatingartifactsgan}.

Recently, researchers have suggested using wavelet transforms to apply a multi-scale frequency approach to GAN image forensics. Specifically, the wavelet transform generates localized frequency decompositions that divide an image into a set of sub-band images that capture distinct frequency ranges, from low-frequency approximations to high-frequency details. In contrast to a single global Fourier spectrum, wavelet transforms can retain spatial localization, which may help isolate specific textured artifacts. Recent studies have formulated wavelet-packet decompositions and presented substantial evidence of distinctions in the wavelet coefficients from real and GAN images \cite{wolter2022wavelet}. They built classifiers on wavelet representations of the images and achieved detection accuracy that matched or exceeded pixel-domain CNNs while employing smaller network architectures.

This research builds upon these insights and investigates a deep learning approach to GAN image detection via wavelet preprocessing. In particular, the proposed method contains a wavelet-based StyleGAN fingerprint detection method with ResNet50 as the classifier. Discrete Wavelet Transform (DWT) is applied using two filters, Haar wavelet and Daubechies wavelet, to transform input images before they are passed into the network. The rationale is that wavelet decomposition will emphasize high-frequency artifacts (e.g., up-sampling patterns, checkerboard noise, and missing camera sensor noise) that distinguish GAN outputs from real images. Thus simplifying the classification task for the neural network.

\section{Data and CNN Architecture}

The dataset contained two primary semantic classes: human-face images and cat images. Within those classes, a balanced set of real and GAN-generated images was acquired to facilitate robust training and evaluation of image authenticity detection models.

\subsection{Real Images Collection}

For Real Face Images, 2,500 images were randomly selected from the Flickr-Faces-HQ (FFHQ) \cite{ffhq_dataset} human faces. Altogether, FFHQ has 70,000 high-definition human face photos with substantial demographic variability. FFHQ has great variability in age, gender, ethnicity, facial expression, and backgrounds, making it a suitable dataset that mimics the reality of diverse human faces. All images were resized to 256×256 for consistency.

\begin{figure}[ht]
    \centering
    \includegraphics[width=3in]{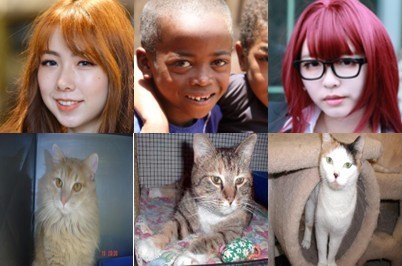}
    \caption{A sample of real images, including human faces (from FFHQ) and cat images (from the Cats vs. Dogs dataset). These images are the true class for the binary classification task. They also provide variety in terms of age, appearance, and background.}
    \label{fig_real_images}
\end{figure}

For Real Cat Images, 2,500 images were randomly selected from the Cats vs Dogs dataset on Kaggle \cite{cats_vs_dogs_kaggle}, which had been filtered to class cat images. This dataset provided a sufficient assortment of natural, real-world cat images to complete the second class of the dataset. Figure \ref{fig_real_images} presents a sample of images taken from the custom-created dataset that shows real cats and human faces.

\subsection{GAN-generated Images Collection}

Fake Face Images were produced by the StyleGAN2 architecture 
\cite{karras2020analyzingimprovingimagequality}. 2,500 images were randomly selected from a publicly available collection of StyleGAN2-generated faces \cite{stylegan2_github}. The faces are very similar to real humans, and distinguishable artifacts are present from the GAN generation process.

\begin{figure}[ht]
    \centering
    \includegraphics[width=3in]{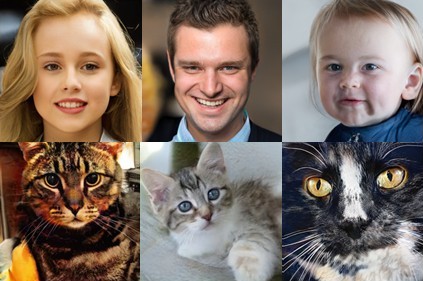}
    \caption{Sample StyleGAN2-generated (fake) images, including synthetic human faces and images of cats. While they are visually realistic, these images contain artifacts (GAN fingerprints) that deep learning models can utilize in the wavelet domain for detection.}
    \label{fig_gan_images}
\end{figure}

A random sample of 2,500 Fake Cat Images was acquired similarly from an existing StyleGAN2-generated synthetic cat image collection to reflect both the distribution and resolutions of real cat images \cite{stylegan2_github}. Figure \ref{fig_gan_images} demonstrates a sample of GAN-generated images showing fake cats and human faces.

No pre- and post-processing processes were used on the images produced by GANs (except resizing when necessary). StyleGAN2 can produce a highly realistic image, but its use of convolutional up-sampling creates fine artifacts that can be undetectable to the human eye but can be detected through spectral analysis in the wavelet domain.

The custom dataset includes both human faces and animals, providing a meaningful yardstick for the potential generalizability of real vs. fake image detectors across various object classes and domains.

\subsection{Dataset Composition}

The overall dataset has an equal distribution of real and synthetic images. 10,000 images were accumulated, divided in a 70:15:15 ratio among the training (7,000), validation (1,500), and testing (1,500) sets, as shown in the Table \ref{tab:dataset_split_detailed}. The test set images were held out entirely during training. The dataset even contained an equal number of cat images and human faces. This 50/50 class balance is maintained to avoid biasing the classifier toward either class. Moreover, data augmentation was performed to enhance training diversity and model robustness. Images are rescaled to normalize pixel values. Random rotations (up to 15°), along with width and height shifts of up to 10\%, and zoom variations of up to 10\%. simulate real-world variations in object positioning and size. Horizontal flipping helps the model learn from mirrored images, while the ``reflect" fill mode preserves edge continuity during transformations.

\begin{table}[ht]
\centering
\caption{Detailed Dataset Composition and Split}
\label{tab:dataset_split_detailed}
\begin{tabular}{|l|c|c|c|c|}
\hline
\textbf{Subset}      & \textbf{Real Images} & \textbf{StyleGAN2 Images} & \textbf{Total Images} \\ \hline
Training Set         & 3,500                & 3,500                     & 7,000                 \\ 
Validation Set       & 750                  & 750                       & 1,500                 \\ 
Test Set            & 750                  & 750                       & 1,500                 \\ \hline
\textbf{Overall Total} & 5,000       &      5,000            &          10,000       \\ \hline
\end{tabular}
\end{table}

\subsection{CNN Architecture}

\begin{figure*}[hbt!]
    \centering
    \includegraphics{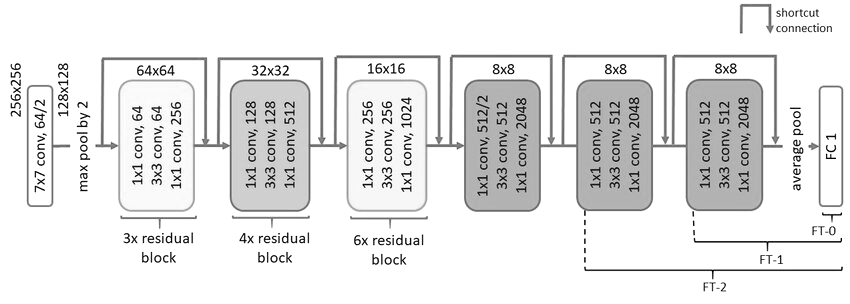}
    \caption{High-level architecture of the ResNet50 deep neural network used for both spatial-domain and wavelet-domain classification tasks.}
    \label{fig_resnet}
\end{figure*}

ResNet50 was utilized as the classifier, which is a 50-layer deep residual network that has demonstrated strong image classification capabilities \cite{he2015deepresiduallearningimage}. Figure \ref{fig_resnet} depicts a high-level architecture of the network. ResNet50 was chosen as it is capable of learning subtle hidden patterns and features in the physical and frequency domains. Notably, the architecture and the hyperparameters were kept consistent throughout all the experiments to ensure an unbiased evaluation between the spatial and frequency domain models.

\section{Methodology}

\subsection{Wavelet Transform}

Wavelet transforms are mathematical techniques for decomposing a signal (or image) into components that are localized in both space (or time) and frequency at multiple scales. In contrast to the Fourier transform, which represents a signal as a combination of global sine and cosine functions, the wavelet transform utilizes small, localized wave-like functions called wavelets. These can effectively capture sharp discontinuities, edges, and transient components. Wavelet transforms for 2D images are referred to as the Discrete Wavelet Transform (DWT), and separate images into four sub-bands of different frequency components across horizontal, vertical, and diagonal directions. Specifically, applying a one-level DWT produces four sub-bands: the low-frequency approximation and three high-frequency detail bands, each containing low to high scale of texture or edge information \cite{agarwal2017analysis, computers13120341, erukude2024identifying, othman2020applications}. This is useful in applications like image compression, denoising, and forensic analysis because it isolates the high-frequency details where synthetic generation artifacts often reside.

\begin{figure}[ht]
    \centering
    \includegraphics[width=2.5in]{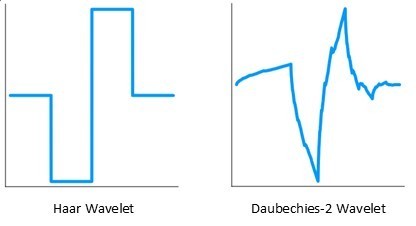}
    \caption{Illustration of the Haar and Daubechies wavelet filters used for image decomposition, highlighting differences in filter shapes.}
    \label{fig_wavelets}
\end{figure}

Among the various wavelet families, Haar and Daubechies wavelets are two widely used types, each with distinct properties as shown in the Figure \ref{fig_wavelets}, and this study utilizes both Haar and Daubechies wavelets to transform images into the wavelet domain.

\subsection{Haar Wavelet}

\begin{figure}[ht]
    \centering
    \includegraphics[width=3.4in]{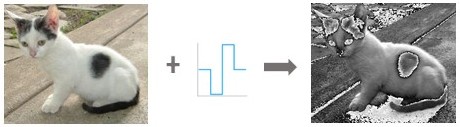}
    \caption{Example of a real cat image before and after applying Haar wavelet.}
    \label{fig_haar_example}
\end{figure}

Haar wavelet is the simplest and earliest wavelet, equivalent to Daubechies-1. Uses a two-tap filter that performs a basic averaging and differencing operation. It is excellent for detecting sharp edges and sudden changes in intensity, but has limited capacity to capture smooth or oscillatory patterns. Figure \ref{fig_haar_example} depicts a real cat image before and after applying the Haar wavelet.

\subsection{Daubechies Wavelet}

\begin{figure}[ht]
    \centering
    \includegraphics[width=3.4in]{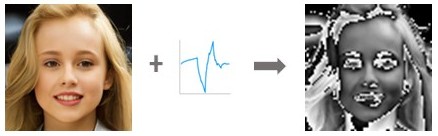}
    \caption{Example of a StyleGAN2-generated human face image before and after applying Daubechies-2 wavelet.}
    \label{fig_db2_example}
\end{figure}

Daubechies wavelets, introduced by Ingrid Daubechies, are a set of orthogonal wavelets defined by their number of vanishing moments, in that they can represent polynomials of a certain order with zero coefficients. For instance, the Daubechies-2 (db2) wavelet has one vanishing moment, has a longer filter length than Haar, and can depict smoother variations and track subtle differences across larger neighborhoods. Hence, Daubechies wavelets can be particularly effective at depicting textures, gradients, and quickly changing complex oscillations. This is particularly important when attempting to identify subtle artifacts left behind from complex GAN models. Figure \ref{fig_db2_example} shows a GAN-generated fake face image before and after applying the Daubechies-2 wavelet using the ``PyWavelets" Python package \cite{lee2019pywavelets}.

\subsection{Training Setup}

All three models, DWT-based (Haar, Daubechies) and spatial, were developed with supervised learning to classify images as ``Real" (class 0) or “Fake” (StyleGAN; class 1). A binary cross-entropy loss function, an Adam optimizer with a learning rate of 0.0001, and a batch size of 32 were employed. Several out-of-the-box callbacks, namely ModelCheckpoint, EarlyStopping, and ReduceLROnPlateau, were utilized in the model training. These callbacks help us save the best-performing model, avoid overfitting, and reduce the learning rate when plateaued. The dataset was split into training, validation, and testing sets, with a ratio of 0.70:0.15:0.15.

The following evaluation metrics were captured during the evaluation phase: classification accuracy, the ROC curve and AUC, and Average Precision, which provides an overall summary of the precision-recall trade-off. These metrics give a full picture of performance. The threshold-agnostic separation (AUC, AP) to the specific fixed threshold measure (accuracy with a threshold of 0.5).

\section{Results and Discussion}

After training, the spatial domain and wavelet domain ResNet50 models were assessed on the 1,500-image test set. Both the DWT-based models significantly outperformed the spatial model on all metrics. The spatial model attained about $\approx 81.5\%$ accuracy on the test set, which indicates it can distinguish StyleGAN2 fakes from real images fairly well (much better than random guessing).

\begin{figure}[H]
\centering
    \includegraphics[width=3.5in]{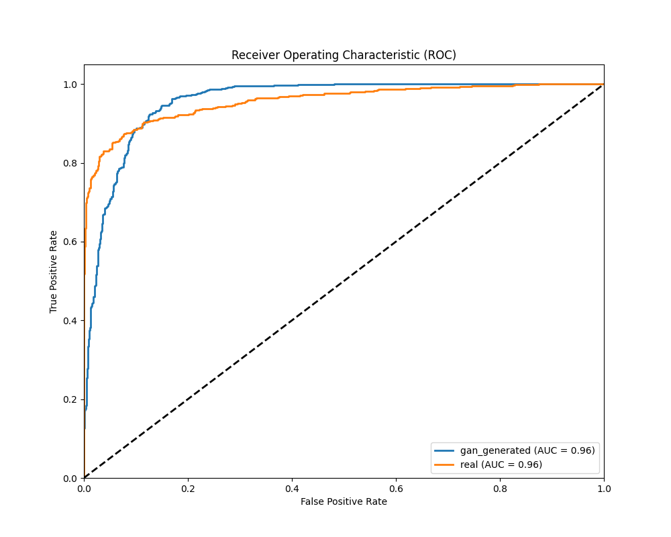}
    \caption{ROC curve for Haar-wavelet ResNet50 model.}
    \label{fig_roc1}
\end{figure}

\begin{figure}[H]
\centering
    \includegraphics[width=3.5in]{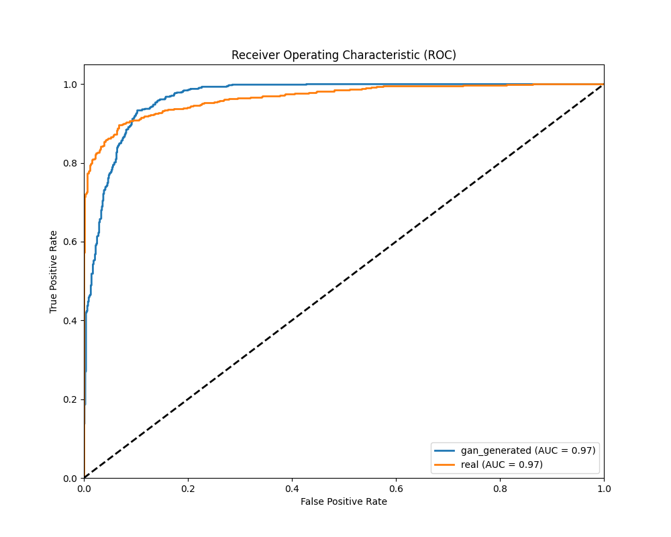}
    \caption{ROC curves for Daubechies-wavelet ResNet50 model, demonstrating superior performance of the Daubechies-wavelet.}
    \label{fig_roc2}
\end{figure}

The Haar wavelet model produced an accuracy of $\approx 93.8\%$ and an AUC of 0.96 (see Figures \ref{fig_roc1} and \ref{fig_roc2}). This equates to 12\% more correct classifications than the spatial model, and is closer than other models to the upper-left (an ideal detector would have AUC = 1.0). The AP of 0.88 indicates a very good relative ranking of positive (fake) vs negative (real) across thresholds; this is a nice improvement over the 0.85 AP from the baseline.

The Daubechies wavelet model indicated slightly improved performance when considered in terms of accuracy ($\approx 95.1\%$), AUC (0.97), and AP (0.89). Essentially, it detected 95 out of 100 StyleGAN2 images correctly on average. The performance gap between the Daubechies-2 model and the Haar model is smaller (roughly 1.3\%), but consistently in favor of Daubechies across all metrics. The specific model metrics are summarized in Table \ref{tab_performance}.

\begin{table}[H]
\centering
\caption{Performance comparison of spatial and wavelet-based ResNet50 models on the 1,500-image test set.}
\label{tab_performance}
\begin{tabular}{|p{2cm}|p{2cm}|p{2cm}|p{1cm}|}
\hline
\textbf{ResNet50 Model} & \textbf{Accuracy (\%)} & \textbf{F1 Score} & \textbf{AUC} \\
\hline
Spatial      & 81.5    & 0.802      & 0.85   \\     
Haar         & 93.8    & 0.872      & 0.96     \\    
Daubechies-2   & 95.1    & 0.886      & 0.97 \\
\hline
\end{tabular}
\end{table}

\begin{figure}[H]
\centering
    \includegraphics[width=3.5in]{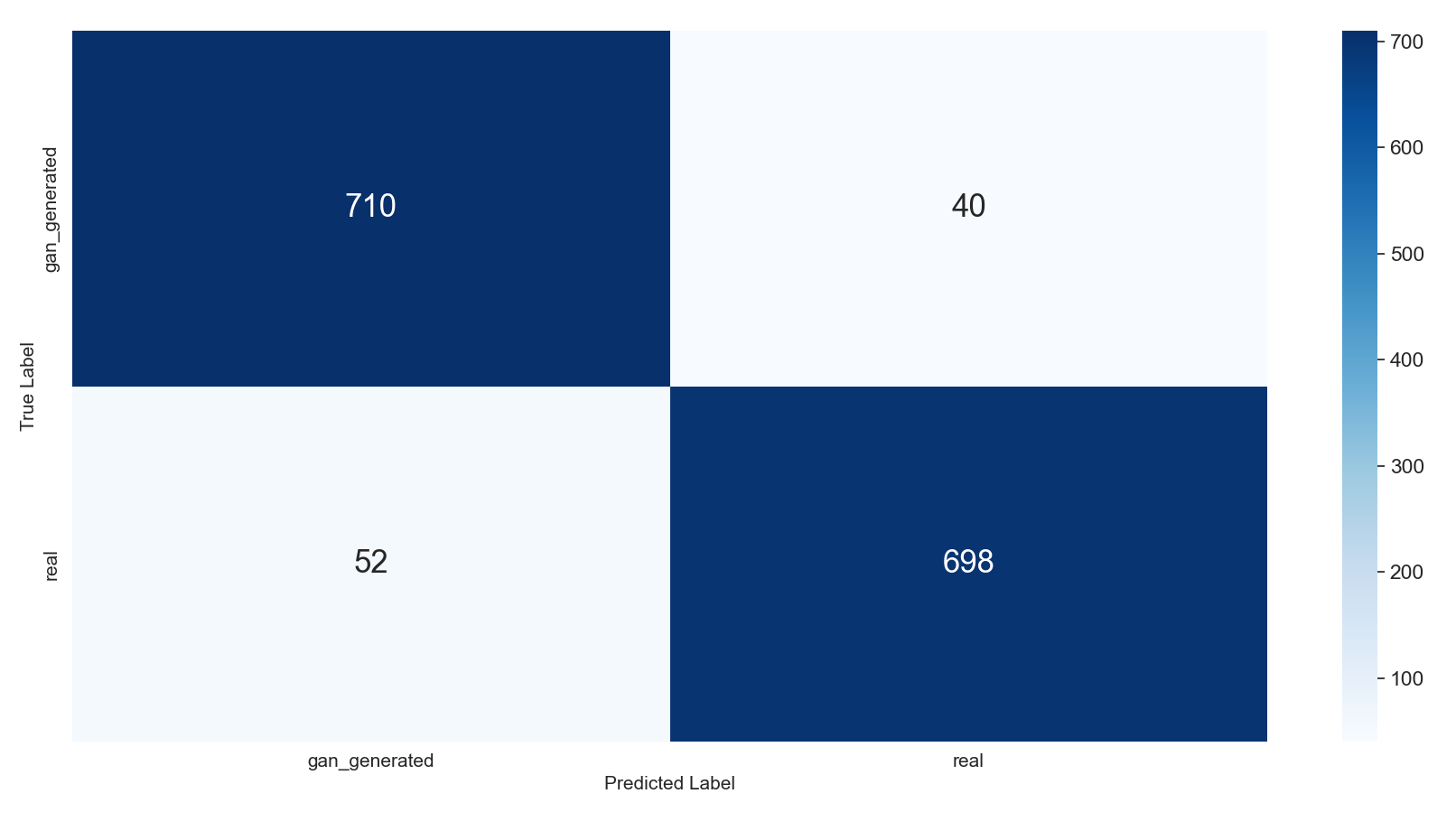}
    \caption{Confusion matrix for ResNet50 model trained on Haar-transformed test set.}
    \label{fig_cm1}
\end{figure}

\begin{figure}[H]
\centering
    \includegraphics[width=3.5in]{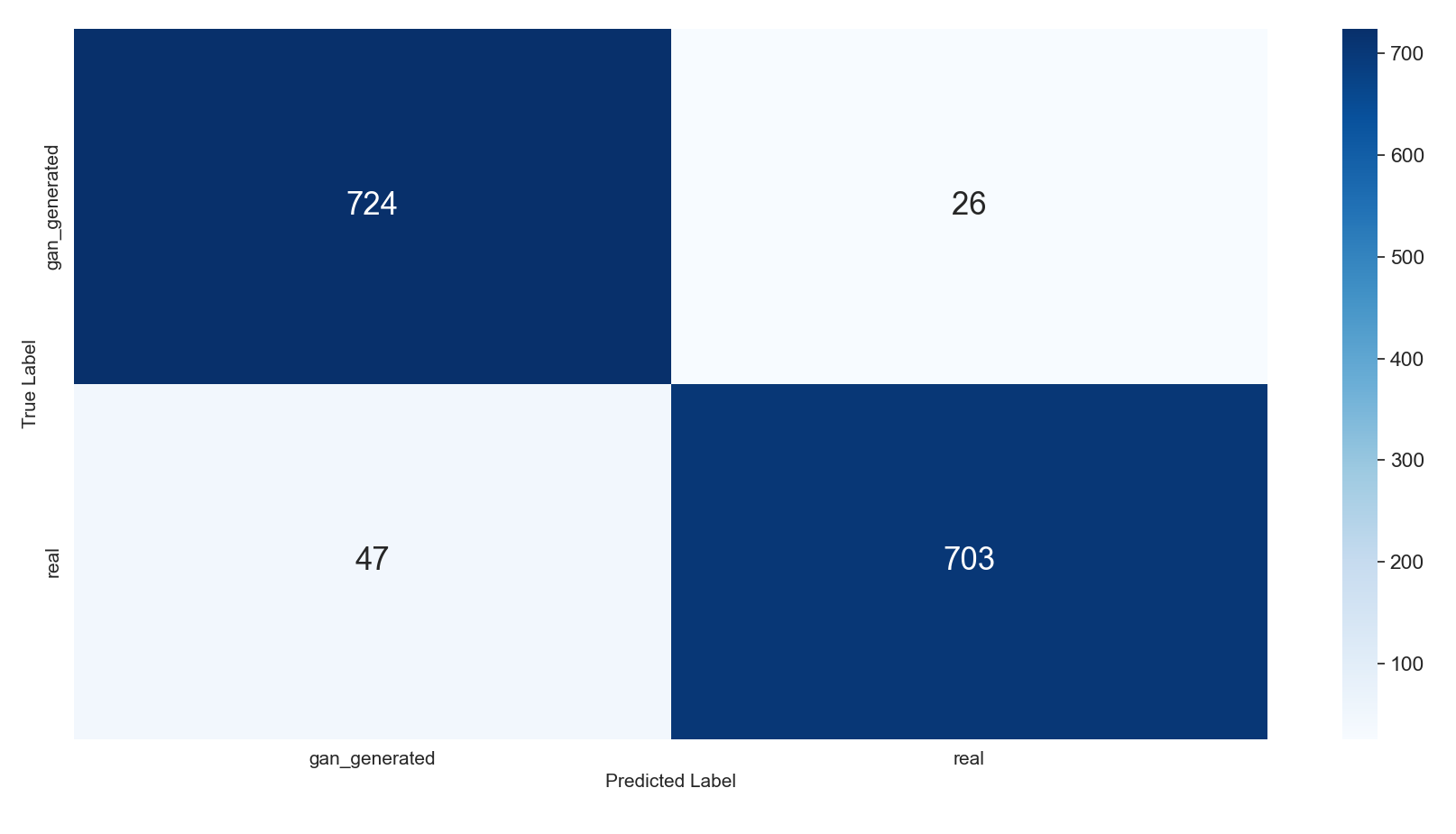}
    \caption{Confusion matrix for ResNet50 model trained on Daubechies-transformed test set, showing improved true positive and true negative rates.}
    \label{fig_cm2}
\end{figure}

Notably, Table \ref{tab_performance} also shows that the wavelet-domain classifier had roughly a 13.6\% increase in accuracy over the spatial classifier. This difference is significant, especially considering the models were trained on the same architecture, which tells us the wavelet-domain input is providing a more useful signal for GAN detection. Figure \ref{fig_cm1} and \ref{fig_cm2} illustrate the confusion matrices of ResNet50 models trained on Spatial, Haar, and Daubechies transformed data. The DWT-based models achieve higher true positive and true negative rates with fewer misclassifications. This highlights that wavelet-domain analysis, especially the Daubechies-2 wavelet, provides a stronger and more reliable signal for distinguishing StyleGAN2 fingerprints compared to raw spatial features.

The statistically significant improvement in the performance of wavelet-transformed inputs, with the Daubechies wavelet in particular, indicates that GAN-generated images include artifacts that are not easily viewed in the spatial domain, but rather gestural artifacts at the frequency level. While Haar is focused on sharp or abrupt transitions in intensity, the Daubechies wavelet is more concerned with oscillation and captures smoother transitions, which may reveal asymmetries in the synthesis of texture and background gradients. This was an interesting point considering that a qualitative analysis of the false positives produced by the spatial model showed many of the images did not feature much geometrical distortion. Instead, these images featured unnatural micro-textures, a form of distortion that would exist in the high-frequency DWT sub-bands. Therefore, frequency-domain decompositions may prove to be a potent forensic tool against advanced GANs.

\section{Conclusion}

This paper presented a study of wavelet-based GAN fingerprint detection using ResNet50, which showed that discrete wavelet transforms (DWT) can greatly improve a deep neural network's ability to distinguish between GAN-generated images and real photographs. Two DWT preprocessing methods (Haar and Daubechies wavelets) were implemented and compared to a standard spatial input, with all experiments using a ResNet50 classifier. Transforming images into the wavelet domain with 2D DWT allowed exposing hidden periodic artifacts (``GAN fingerprints”) that a deep ResNet50 classifier could learn to recognize with high confidence. Haar wavelet improved accuracy, AUC, and average precision over the baseline, and Daubechies with a more complex wavelet performed best in most metrics (e.g., ~95.1\% test accuracy and 0.97 AUC). These results confirm that wavelet decomposition uncovers forensic details, or GAN fingerprints, that are not as evident in raw pixel data. It was also found that using a more complex wavelet (Daubechies-2) provided a slight edge over the simple Haar wavelet, suggesting that capturing more nuanced frequency patterns can further boost detection. The ResNet50, a powerful deep feature extractor, leveraged these wavelet features effectively to distinguish real vs. fake. From a broader perspective, this study highlights a successful integration of signal processing techniques with modern deep learning for image forensics. It contributes to the increasing body of evidence leaning in favor of frequency-domain (Fourier or wavelets) analyses being advantageous for distinguishing AI-generated content. The wavelet + ResNet50 method generated in this paper could be further expanded or deployed into usable fake image detection systems. As GAN-generated media becomes increasingly used and realistic, the need for hybrid techniques that draw on both human-understandable transformations and machine learning could be substantial to maintain the integrity of digital imagery. Future forensic applications could learn multi-resolution and multi-domain features to develop capabilities to keep pace with an arms race effectively against generative models, by investigating Cross-GAN Generalization and robustness to post-processing.


\section*{ACKNOWLEDGMENT}

The authors thank the contributors of the public datasets used in this study. The complete Python code has been made available at: \url{https://github.com/SaiTeja-Erukude/gan-fingerprint-detection-dwt}.

\bibliographystyle{IEEEtran}
\bibliography{main}

\end{document}